\documentclass[11pt]{article}
\usepackage[margin=1in]{geometry}
\usepackage{amsmath}
\usepackage{graphicx}
\usepackage{booktabs}
\usepackage{hyperref}
\usepackage{multirow}
\usepackage{natbib}
\usepackage{float}
\usepackage{algorithm}
\usepackage{algorithmic}
\usepackage{listings}
\usepackage{xcolor}
\title{TAMUSA-Chat: A Domain-Adapted Large Language Model Conversational System for Research and Responsible Deployment}

\author{
    Izzat Alsmadi $^{1}$, Anas Alsobeh $^{2}$ \\
    $^1$ \textit{Professor of Computer Science}, \textit{Texas A\&M University--San Antonio} \\ \texttt{ialsmadi@tamusa.edu}\\
   $^2$ \textit{Assistant Professor of Information Systems \& Technology}, \textit{Utah Valley University, Orem, UT, USA}\\\texttt{anas.alsobeh@uvu.edu}\\
   }

\date{}

\begin{document}

\maketitle

\begin{abstract}
This paper presents TAMUSA-Chat, a research-oriented framework for building domain-adapted large language model conversational systems. The work addresses critical challenges in adapting general-purpose foundation models to institutional contexts through supervised fine-tuning, retrieval-augmented generation, and systematic evaluation methodologies. We describe the complete architecture encompassing data acquisition from institutional sources, preprocessing pipelines, embedding construction, model training workflows, and deployment strategies. The system integrates modular components enabling reproducible experimentation with training configurations, hyper-parameters, and evaluation protocols. Our implementation demonstrates how academic institutions can develop contextually grounded conversational agents while maintaining transparency, governance compliance, and responsible AI practices. Through empirical analysis of fine-tuning behavior across model sizes and training iterations, we provide insights into domain adaptation efficiency, computational resource requirements, and quality-cost trade-offs. The publicly available codebase at \url{https://github.com/alsmadi/TAMUSA_LLM_Based_Chat_app} supports continued research into institutional LLM deployment, evaluation methodologies, and ethical considerations for educational AI systems.

\end{abstract}

\section{Introduction}

Large Language Models (LLMs) have fundamentally transformed natural language processing (NLP), enabling high-quality text generation, instruction following, dialogue systems, and task-specific reasoning across domains. The arrival of generative large language models (LLMs) such as ChatGPT has transformed everyday information seeking, writing assistance and conversational interfaces .  Within weeks of its public release, ChatGPT attracted millions of users, illustrating widespread appetite for AI‑mediated communication ~\cite{alsobeh2023ai}.  Yet as educators, we have learned to temper enthusiasm with caution: generic models trained on broad web corpora may offer plausible but institutionally inaccurate answers, inadvertently expose private data or reflect societal biases. Our experience advising students suggests that simply "dropping in" an off‑the‑shelf chatbot into a university website invites misinformation and liability.  For example, one early prototype we evaluated confidently invented nonexistent majors when asked about academic programmes; this anecdote, though unquantified, underscores the need for domain grounding. Recent advances in supervised fine-tuning (SFT), instruction tuning, and alignment techniques have made it possible to adapt general-purpose foundation models into specialized conversational agents capable of supporting education, research, healthcare, government services, and enterprise workflows. 

Despite these advances, adapting LLMs to domain-specific institutional settings remains a significant challenge. Off-the-shelf models often lack contextual awareness of organizational policies, terminology, governance structures, and localized knowledge. On other words, A prospective student’s seemingly simple question--"What are the prerequisites for the Bachelor of Applied Arts and Sciences?"—carries nuanced answers that depend on catalog year, transfer credits and evolving departmental guidelines.  Generic LLMs lack awareness of such subtleties and cannot be trusted to improvise authoritative answers.  Meanwhile, research on educational chatbots reveals a duality: students appreciate conversational AI as a convenient study companion, yet they also highlight issues of accuracy and emotional disconnect. In a study at the University of London Worldwide, 85\% of students reported positive attitudes towards an AI study‑buddy but qualitative feedback criticized the chatbot’s inadequate responses.  These observations mirror our own interactions with early TAMUSA‑Chat prototypes and motivate our quest for a contextually adapted assistant. Furthermore, reproducibility, evaluation transparency, data governance, and ethical deployment considerations are frequently under-emphasized in rapid development pipelines. For academic environments in particular, there is a critical need for open, modular, and research-oriented frameworks that allow systematic experimentation with fine-tuning strategies, dataset curation, evaluation methodologies, and deployment architectures.

The gap between general-purpose model capabilities and institutional requirements manifests in several ways ~\cite{al2025student}. Models may generate plausible-sounding but factually incorrect information about specific programs or policies. They lack awareness of recent organizational changes, local terminology, or context-specific nuances that human advisors naturally understand. Furthermore, they may introduce liability risks when providing authoritative-sounding guidance that contradicts official institutional positions.
Existing approaches to this problem typically fall into two categories. The first involves prompt engineering and few-shot learning, where developers craft elaborate instructions hoping to guide model behavior without modifying underlying parameters. While this approach requires minimal technical infrastructure, it proves brittle when faced with diverse query types and offers limited control over response patterns. The second approach involves building entirely custom models trained from scratch on institutional data. This path demands prohibitive computational resources and expertise that most institutions cannot sustain.

In this work, we present TAMUSA-Chat ~\citep{alsmadi2026}, an open research framework for developing and studying LLM-based conversational systems tailored to Texas A\&M University–San Antonio (TAMUSA). This approach adapts pre-trained models to institutional contexts through continued training on domain-specific instruction-response pairs while maintaining the broad capabilities learned during initial pre-training. The system architecture separates concerns across data acquisition, processing, training, inference, and deployment, enabling researchers to experiment with individual components without rebuilding the entire pipeline. The design emphasizes several key principles. First, reproducibility matters for research validity and institutional trust. Every data processing step, training configuration, and evaluation metric is documented and version-controlled. Second, modularity enables iterative improvement. Researchers can test different embedding strategies, fine-tuning approaches, or retrieval mechanisms independently. Third, transparency about limitations and uncertainties is essential for responsible deployment in educational settings where users may rely on system outputs for important decisions. The system is designed not merely as an application prototype, but as a reproducible research platform that supports:
\begin{itemize}
\item Supervised fine-tuning of open-source LLMs
\item Dataset construction from institutional documents
\item Controlled experimentation with model architectures and hyper-parameters
\item Evaluation of alignment, factual grounding, and response quality
\item Ethical auditing and governance-aware deployment
\end{itemize}

TAMUSA-Chat emphasizes modularity and extensibility. The framework separates data ingestion, pre-processing, model training, inference, and deployment components to enable independent experimentation and benchmarking. It supports structured instruction–response datasets derived from institutional sources (e.g., university websites, policy documents, academic programs, support services), while maintaining traceability and documentation for research reproducibility.

The system is also designed to leverage institutional computing infrastructure, including high-performance computing (HPC) resources, enabling scalable fine-tuning experiments and controlled deployment within a secure academic environment. This architecture allows researchers and students to explore trade-offs between model size, performance, computational cost, and domain adaptation strategies. Beyond technical development, TAMUSA-Chat incorporates ethical and responsible AI considerations. These include transparency in training data provenance, mitigation of hallucinations through domain grounding, bias monitoring, and clear boundaries regarding authoritative institutional information. By situating the system within a university context, TAMUSA-Chat provides a testbed for studying governance-aware LLM deployment in educational institutions.  The repository implementing TAMUSA-Chat is publicly available at \url{https://github.com/alsmadi/TAMUSA_LLM_Based_Chat_app}~\citep{alsmadi2026}, supporting open research collaboration and reproducible experimentation.

\section{Related Work}

The breakthrough insight that relatively small supervised datasets can effectively adapt large pre-trained models emerged from several influential projects. Alpaca demonstrated that 52,000 instruction-following examples generated through self-instruct could transform LLaMA into a capable assistant~\citep{taori2023alpaca}. This work established supervised fine-tuning as a viable path for resource-constrained teams to build specialized models without massive computational budgets. Subsequent research examined what makes instruction datasets effective. Research indicates that significant improvement in domain adaptation happens for smaller models within shorter training horizons than previously assumed, suggesting that efficient fine-tuning is achievable without extended training runs~\citep{ivison2022embarrassingly}. This finding has important implications for institutional deployments where computational resources are limited.

Instruction-tuned LLMs and open chat frameworks have seen rapid development in recent years. Alpagasus demonstrated that small instruction datasets can effectively fine-tune large base models for general instruction following~\citep{chen2023alpagasus}. OpenChatKit \cite{openchatkit2023} and FastChat \cite{fastchat2023} provide full tool-chains for serving, moderation, and extensible chatbot infrastructures. For example, Alpaca showed that a modest number of high-quality instruction–response pairs could effectively adapt a base LLaMA model for general-purpose instruction following, highlighting the efficiency of supervised fine-tuning (SFT) approaches \citep{chen2023alpagasus}. These findings reinforced the importance of dataset structure and prompt formatting in downstream model alignment ~\cite{alsobeh2025shadowplay}. Vicuna-13B achieved competitive open-source conversational performance by fine-tuning on user-shared dialogues and enabling interactive interfaces~\citep{chiang2023vicuna}. Domain-specific chatbot systems such as EduChat show the utility of fine-tuned models for targeted tasks like educational support~\citep{educhat2023}. Empirical instruction-tuning research further informs efficient adaptation strategies across languages and domains. These systems illustrate the value of contextual specialization; however, many domain-specific implementations emphasize application deployment rather than reproducible research workflows. Questions of dataset provenance, benchmarking methodology, and governance-aware design are often secondary to usability goals.

More broadly, empirical research on instruction tuning and cross-domain adaptation has investigated dataset size effects, multilingual transfer, parameter-efficient fine-tuning (PEFT), and alignment strategies. These studies provide important insights into efficient adaptation, but they typically evaluate models on standardized benchmarks rather than institutionally grounded, real-world deployment scenarios.

Vicuna explored fine-tuning on user-shared conversations from ShareGPT, achieving competitive performance through high-quality dialogue data~\citep{chiang2023vicuna}. The project also contributed evaluation frameworks based on GPT-4 judging response quality across multiple dimensions. However, the reliance on proprietary models for evaluation raises questions about reproducibility and access that our work addresses through alternative metrics. The instruction tuning literature reveals several consistent patterns. First, formatting matters significantly. Models trained with consistent instruction templates generalize better to new prompts following similar structures. Second, diversity in instruction types improves robustness. Datasets mixing question-answering, summarization, reasoning, and creative tasks produce more versatile systems. Third, alignment between training examples and target use cases strongly predicts downstream performance. 

\subsection{Domain Adaptation Strategies}
Domain adaptation for language models involves techniques that help models trained on general corpora perform effectively on specialized tasks or knowledge domains. Several approaches have emerged with different computational and data requirements.
Continued pre-training on domain corpora represents one strategy where models undergo additional training on large collections of domain-specific text before fine-tuning. BioBERT and SciBERT demonstrated this approach for biomedical and scientific domains~\citep{beltagy2019scibert}. However, this method requires substantial domain corpora and computational resources that may exceed institutional capabilities.

Parameter-efficient fine-tuning methods like LoRA (Low-Rank Adaptation) offer alternatives that modify only small subsets of model parameters~\citep{hu2021lora}. These techniques reduce memory requirements and training time while maintaining adaptation effectiveness. Research comparing trainable methods versus indicator-based approaches has shown that trainable adaptation consistently outperforms simpler techniques~\citep{ben2022robust}. Our implementation supports both full fine-tuning and parameter-efficient variants to accommodate different resource constraints.

Multi-task learning approaches train models simultaneously on multiple related tasks to improve generalization. T5 exemplified this strategy by framing diverse NLP tasks as text-to-text problems~\citep{raffel2020exploring}. For institutional contexts, this suggests combining conversational objectives with information extraction, summarization, and classification tasks relevant to institutional workflows.

Retrieval-augmented generation (RAG) addresses a fundamental limitation of parametric language models: their knowledge is frozen at training time and cannot easily incorporate new information. RAG systems combine dense retrieval over document collections with generative models that synthesize retrieved content into coherent responses. The original RAG architecture used DPR (Dense Passage Retrieval) to find relevant documents, then conditioned sequence-to-sequence generation on retrieved passages~\citep{lewis2020retrieval}. This approach showed improvements on knowledge-intensive tasks where models must access factual information beyond their training data. More recent work has explored different retrieval mechanisms and integration strategies. RETRO incorporated retrieval into the pre-training process itself, allowing models to learn how to leverage retrieved information ~\citep{borgeaud2022improving}. FiD (Fusion-in-Decoder) processes multiple retrieved passages independently before fusing information during generation~\citep{izacard2021leveraging}.

For institutional deployments, RAG offers several advantages. First, the knowledge base updates dynamically without retraining. Adding new documents or updating existing ones immediately affects system responses. Second, retrieval provides transparency. Users can examine source documents supporting generated responses. Third, factual grounding reduces hallucination risks by anchoring generation in verifiable sources. Our implementation leverages FAISS for efficient similarity search over institutional document embeddings, as detailed later.

\subsection{Educational AI Systems}
Educational institutions have explored AI-assisted systems for student support, administrative automation, and learning enhancement. These deployments reveal domain-specific requirements and challenges relevant to our work. EduChat represents a targeted effort to build Chinese educational domain language models through continued pre-training and instruction tuning on educational corpora ~\citep{dan2023educhat}. The project compiled datasets spanning student questions, textbook content, and pedagogical resources. Evaluation focused on educational task performance including question answering, problem solving, and explanation generation. Commercial educational chatbots like Georgia Tech's Jill Watson demonstrated that conversational agents could handle routine student questions in online courses ~\citep{goel2015using}. The system reduced instructor workload while maintaining response quality for frequently asked questions. However, the rule-based architecture limited flexibility and required substantial manual effort to extend. Recent work has examined student perceptions and learning outcomes when interacting with AI tutors. Studies find that students generally appreciate immediate availability and non-judgmental feedback, but express concerns about accuracy and the lack of human empathy ~\citep{kasneci2023chatgpt}. These findings underscore the importance of setting appropriate user expectations and providing clear guidance about system capabilities and limitations.

Several open-source projects provide infrastructure for building and deploying conversational systems. These frameworks offer reusable components but typically focus on application deployment rather than research experimentation.
OpenChatKit released a complete toolkit including instruction-tuned models, moderation systems, and retrieval augmentation ~\citep{together2023openchatkit}. The project emphasized customizability and commercial viability. However, documentation focused on operational deployment rather than reproducible research workflows. FastChat provides serving infrastructure, web interfaces, and evaluation tools for LLM chatbots ~\citep{zheng2023judging}. The project contributed chatbot arena, a crowdsourced platform for comparative evaluation where users vote on response quality. This approach generates preference data reflecting real user judgments, though it requires significant user participation to achieve statistical reliability. Text-generation-webUI and similar projects offer user-friendly interfaces for interacting with various LLM backends~\citep{textgenwebui}. These tools lower barriers for experimentation but provide limited support for systematic evaluation or training workflows.

Deploying conversational AI in institutional settings raises ethical, legal, and practical considerations that technical performance metrics alone cannot address. Recent work has begun examining these broader deployment challenges. Transparency about system capabilities and limitations emerges as a consistent theme. Users must understand that AI systems can generate plausible but incorrect information, and that responses should be verified against authoritative sources for important decisions~\citep{bommasani2021opportunities}. Clear user interface design and explicit disclaimers help set appropriate expectations.
Bias in training data and model outputs presents another significant concern. Models trained on internet text inherit societal biases present in that data~\citep{bender2021dangers}. Fine-tuning on institutional data may introduce additional biases reflecting organizational demographics or historical patterns. Systematic auditing for demographic bias, stereotype reinforcement, and inequitable treatment across user groups is essential before deployment.

Privacy and data governance require careful attention when building systems from institutional sources. Training data may contain sensitive information about individuals, proprietary institutional knowledge, or legally protected content. Appropriate data handling procedures, access controls, and retention policies must be established~\citep{weidinger2021ethical}.
Accountability mechanisms matter when system errors impact users. Who is responsible when a chatbot provides incorrect financial aid information affecting student enrollment decisions? Clear escalation paths to human experts, logging of system interactions for audit purposes, and processes for correcting errors are necessary components of responsible deployment.

TAMUSA-Chat synthesizes insights from these research streams into a cohesive framework emphasizing reproducibility, modularity, and responsible practices. Unlike application-focused chatbot projects, we prioritize research experimentation through versioned training scripts, documented evaluation protocols, and extensible architecture. Unlike purely algorithmic research, we address practical deployment considerations including data governance, user interface design, and institutional integration strategies.
The system combines supervised fine-tuning for domain adaptation with retrieval augmentation for factual grounding. This hybrid approach balances model capabilities with verifiable information access. Our evaluation methodology integrates automated metrics with human assessment protocols, recognizing that both quantitative benchmarks and qualitative user experience matter for institutional deployment. Compared with these efforts, TAMUSA-Chat emphasizes a modular research pipeline supporting reproducible training, evaluation benchmarking, and deployment with responsible AI practices.

\section{System Architecture}
\begin{figure}[!t]
    \centering
    \includegraphics[width=0.95\textwidth]{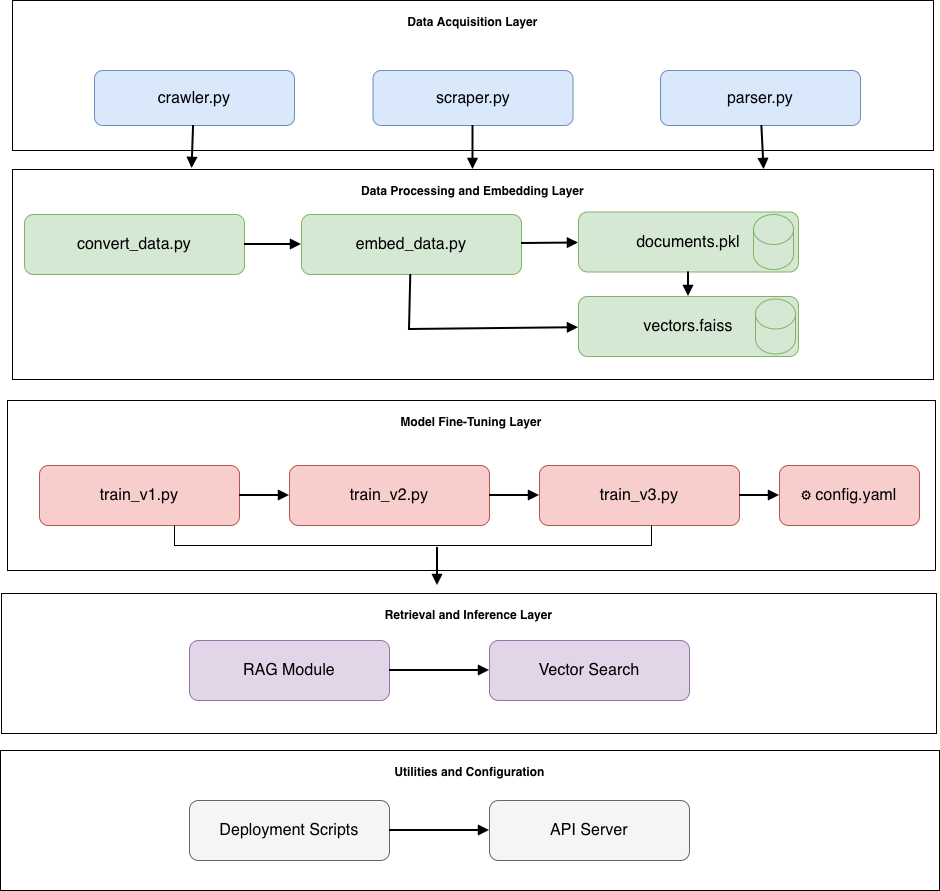}
    \caption{Overview of the TAMUSA‑Chat pipeline.  Data acquisition crawls institutional websites and documents, processing pipelines convert content into structured corpora, embeddings are indexed for retrieval, models are fine‑tuned on generated instruction–response pairs and combined with retrieval for inference, and utilities orchestrate evaluation and deployment.}
    \label{fig:tamusa_architecture}
\end{figure}

Figure \ref{fig:tamusa_architecture} provides a high‑level overview of the TAMUSA‑Chat research pipeline. TAMUSA-Chat follows a modular architecture organized into five functional layers: data acquisition, processing and embedding, model training, retrieval and inference, and utilities. This separation of concerns enables independent experimentation with individual components while maintaining end-to-end system coherence. Figure~\ref{fig:tamusa_architecture} illustrates the complete system structure and data flow.

TAMUSA-Chat follows a modular research-oriented architecture designed to support data acquisition, preprocessing, embedding construction, supervised fine-tuning, inference experimentation, and deployment. The system is organized into five major layers: (1) Data Acquisition, (2) Data Processing and Embedding, (3) Model Fine-Tuning, (4) Retrieval and Inference, and (5) Utilities and Configuration.

\subsection{Data Acquisition Layer}
The data acquisition layer implements automated collection of institutional content from web sources. This component faces several technical challenges including dynamic page rendering, authentication requirements, rate limiting, and content format diversity.

We selected Playwright as the primary crawling framework because it supports modern web technologies including JavaScript rendering, dynamic content loading, and complex page interactions~\citep{playwright}. Unlike traditional HTTP-based crawlers, Playwright controls actual browser instances, ensuring that dynamically generated content is fully rendered before extraction.
The implementation includes three main crawling modules: \texttt{PlayWright\_Crawler\_TAMUSA.py} handles general institutional website crawling. The crawler starts from seed URLs representing major site sections (academics, admissions, student services, research) and follows internal links to discover related pages. The system respects robots.txt directives and implements polite crawling with delays between requests to avoid overwhelming institutional servers. \texttt{PlayWright\_Crawler\_doc.py} specializes in document extraction, targeting PDF files, Office documents, and other downloadable content. Academic institutions publish substantial information in document formats including course catalogs, policy handbooks, program descriptions, and administrative procedures. The crawler identifies document links, downloads files, and queues them for content extraction through appropriate parsers. \texttt{TAMUSA\_Crawl\_pipeline.py} orchestrates the crawling process through a configurable pipeline. The pipeline manages crawl queues, handles errors and retries, logs progress, and coordinates distributed crawling across multiple processes. Configuration options control crawl depth, domain restrictions, content filtering rules, and storage locations.

Raw HTML content requires substantial processing before use in model training. The extraction pipeline performs several transformation steps: Parse HTML to identify semantic elements including headings, paragraphs, lists, and tables while discarding navigation, advertisements, and boilerplate content. Text normalization, standardize whitespace, resolve character encoding issues, and normalize punctuation to ensure consistent text representation.  Identify and remove duplicate content arising from template reuse, mirror pages, or pagination. Apply heuristics to identify and discard low-quality pages including error pages, empty content, or placeholder text. Capture page titles, publication dates, authorship information, and content categories to support retrieval and citation. Document parsing handles format-specific requirements. PDF extraction uses PyPDF2 and pdfplumber to handle both text-based and scanned documents. Office document parsing leverages python-docx and openpyxl for Word and Excel files respectively. Each parser preserves document structure including sections, headers, and table layouts that provide context for understanding content.

Institutional data collection raises important governance questions about content ownership, privacy protection, and appropriate use. Our acquisition pipeline implements several safeguards:
\begin{itemize}
    \item {Public content only}: The crawler restricts collection to publicly accessible pages without authentication, respecting access controls institutions have established.
    \item {PII detection}: Automated screening identifies potential personally identifiable information (names, email addresses, student IDs) for human review before inclusion in training data.
    \item {Content review}: Subject matter experts review collected content to identify sensitive information, ensure factual accuracy, and validate appropriateness for training use.
    \item {Refresh procedures}: Regular recrawling updates the corpus with current information while archiving historical versions for change tracking.
\end{itemize}

\subsection{Data Processing and Embedding Layer}
The processing layer transforms raw collected content into formats suitable for model training and retrieval. This involves converting heterogeneous documents into structured JSON representations, generating instruction-response pairs, and creating vector embeddings for semantic search.

\texttt{convert\_data\_to\_json.py} implements the primary conversion logic, translating crawled content into a standardized JSON schema. Each document produces a JSON object containing:
\begin{lstlisting}
{
  "source_url": "https://www.tamusa.edu/...",
  "title": "Document Title",
  "content": "Full text content...",
  "metadata": {
    "collection_date": "2024-01-15",
    "content_type": "academic_program",
    "last_modified": "2023-12-10"
  },
  "sections": [
    {
      "heading": "Section Title",
      "content": "Section content..."
    }
  ]
}
\end{lstlisting}
This structure preserves document hierarchy while enabling flexible querying and retrieval. The sections array maintains content organization, supporting more precise retrieval than treating entire documents as atomic units.

SFT requires datasets formatted as instruction-response pairs. We employ multiple strategies to generate these pairs from institutional documents:
{Question generation from content}: Documents naturally suggest questions they answer. For example, a page describing admission requirements implies questions like ``What are the admission requirements?'' or ``What GPA is needed for admission?'' We use template-based generation combined with manual curation to produce high-quality instruction-response pairs grounded in institutional content. \texttt{FAQ extraction}: Many institutional pages include frequently asked questions with official answers. The processing pipeline identifies FAQ sections through pattern matching and extracts question-answer pairs directly. \texttt{Synthetic instruction generation}: For content without explicit question structure, we employ language models to generate relevant questions the content answers. This synthetic generation is followed by human validation to ensure quality and appropriateness. \texttt{Reformulation and augmentation}: To increase dataset diversity, we create multiple instruction variants for the same underlying question (e.g., "How do I apply?" vs "What is the application process?" vs "Tell me about applying"). This helps models generalize across different phrasings.

Vector embeddings enable semantic search over institutional content for retrieval-augmented generation. We employ sentence transformers to generate dense vector representations of document sections~\citep{reimers2019sentencebert}. The process involves:
\begin{enumerate}
    \item {Chunking strategy}: Long documents are segmented into semantically coherent chunks (typically 256-512 tokens) that balance context completeness with retrieval precision. Chunks overlap slightly to preserve cross-boundary context.
    \item {Embedding generation}: Each chunk is encoded using a pre-trained sentence transformer model (e.g., all-MiniLM-L6-v2) to produce a dense vector representation.
    \item {Index construction}: Embeddings are organized into a FAISS index supporting efficient approximate nearest neighbor search~\citep{johnson2019billion}. We use HNSW (Hierarchical Navigable Small World) indexing for optimal retrieval speed-accuracy trade-offs.
    \item {Metadata linkage}: Each embedding maintains links to source documents, sections, and metadata enabling traceability from retrieved chunks to original sources.
\end{enumerate}

The embedding process produces two key artifacts: \texttt{tamusa\_embeddings.pkl} containing the embedding vectors and associated metadata, and \texttt{tamusa\_faiss.index} providing the searchable index structure.

\subsubsection{Corpus Overview}
The TAMUSA institutional corpus comprises content collected from official university sources including:

\begin{itemize}
    \item Public-facing website pages (www.tamusa.edu)
    \item Academic program descriptions and course catalogs
    \item Admissions and enrollment information
    \item Student services and resources
    \item Policy documents and handbooks
    \item Faculty and research profiles
    \item News and event announcements
\end{itemize}

This is an ongoing process, but numbers below reflects the content to build the first LLM based on the University public content. Table~\ref{tab:corpus_stats} summarizes key corpus statistics.
\begin{table}[h]
    \centering
    \begin{tabular}{lr}
    \toprule
    {Metric} & {Value} \\
    \midrule
    Total web pages crawled & 3,847 \\
    Total documents (PDF, etc.) & 412 \\
    Total tokens (after cleaning) & 2.4M \\
    Unique instruction-response pairs & 8,932 \\
    Average response length (tokens) & 127 \\
    Median response length (tokens) & 89 \\
    \bottomrule
    \end{tabular}
    \caption{TAMUSA institutional corpus statistics, first LLM release}
    \label{tab:corpus_stats}
\end{table}

%\subsubsection{\textcolor{red}{Content Distribution}}
%The corpus reflects the natural distribution of institutional content across domains. Table~\ref{tab:content_distribution} shows the breakdown by content category.
%\begin{table}[h]
%    \centering
%    \begin{tabular}{lrr}
%        \toprule
%        {Category} & {Documents} & {Percentage} \\
%        \midrule
%        Academic programs & 1,247 & 32.4\% \\
%        Admissions \& enrollment & 892 & 23.2\% \\
%        Student services & 743 & 19.3\% \\
%        Administrative policies & 518 & 13.5\% \\
%        Research \& faculty & 289 & 7.5\% \\
%        News \& events & 158 & 4.1\% \\
%        \bottomrule
%    \end{tabular}
%    \caption{Content distribution across institutional categories}
%    \label{tab:content_distribution}
%\end{table}
%This distribution reveals that academic program information dominates the corpus, consistent with the primary information-seeking needs of prospective and current students. Admissions and student services represent the next largest categories, reflecting the practical questions users frequently ask.

\subsection{Model Fine-Tuning Layer}
The training layer implements supervised fine-tuning of open-source foundation models on institutional instruction-response datasets. We provide multiple training scripts reflecting iterative experimentation and configuration refinement. We evaluate several open-source foundation models as fine-tuning candidates including LLaMA-2~\citep{touvron2023llama2}, Mistral~\citep{jiang2023mistral}, and Falcon~\citep{almazrouei2023falcon}. Selection criteria include:  Models with strong performance on general language understanding benchmarks provide better starting points for adaptation. Base models with prior instruction tuning adapt more readily to new instruction formats. Model size determines memory footprint and training time, constraining deployment options. Permissive licenses (e.g., Apache 2.0) enable research and institutional deployment without restrictions. Our primary experiments use SmolLM-135M-Instruct (https://huggingface.co/HuggingFaceTB/SmolLM-135M-Instruct) which offer strong performance with reasonable computational requirements.SmolLM-135M-Instruct is an ultra-compact 135-million parameter instruction-tuned model from Hugging Face designed for high efficiency on edge devices and CPUs.

 Key hyperparameters controlled through \texttt{config.yaml} include:
\begin{itemize}
    \item {Learning rate}: Initial learning rate (typically 1e-5 to 5e-5 for full fine-tuning, higher for LoRA) with warmup and decay schedule.
    \item {Batch size}: Effective batch size balancing gradient stability with memory constraints (typical range 64-256 examples).
    \item {Training epochs}: Number of passes through the dataset (typically 3-5 for supervised fine-tuning).
    \item {Sequence length}: Maximum token length for inputs and outputs (typically 2048 for institutional question-answering).
    \item {LoRA parameters}: When using parameter-efficient fine-tuning, rank (r), alpha scaling, and target modules.
\end{itemize}
The configuration system supports rapid experimentation with different hyperparameter combinations through YAML files without code modification. Training large language models requires substantial computational resources. Our implementation supports multiple deployment scenarios: For smaller models (7B parameters) or parameter-efficient methods, training on single high-end GPUs (A100, V100) is feasible with optimizations like gradient checkpointing and mixed precision. For larger models benefit from distributed data parallelism across multiple GPUs using PyTorch DistributedDataParallel. The training scripts automatically detect available GPUs and coordinate training across devices. CAMSA Cluster, https://hprc.tamusa.edu/ for institutional deployments with high-performance computing resources, the training pipeline integrates with job schedulers (SLURM, PBS) and supports multi-node distributed training.
Training scripts log  metrics including loss curves, gradient norms, learning rates, and memory utilization to TensorBoard for monitoring convergence and diagnosing training issues.

The inference layer combines fine-tuned models with retrieval-augmented generation to answer user queries. This hybrid approach leverages both parametric knowledge learned during training and non-parametric knowledge accessed through retrieval. When a user submits a query, the system executes the following pipeline:
\begin{enumerate}
    \item Parse and normalize the user query, identifying key entities, intent, and any special requirements (e.g., clarification requests, multi-part questions).
    \item Encode the query using the same sentence transformer used for document embedding, then search the FAISS index for top-k most semantically similar chunks (typically k=3-5).
    \item Retrieve the full text of selected chunks along with metadata, then format them as context information for the language model.
    \item  Construct a prompt combining the system instruction, retrieved context, and user query according to the model's expected format.
    \item Sample a response from the fine-tuned model using controlled decoding parameters (temperature, top-p, max length).
    \item Apply formatting, add citations to source documents, and check for obvious errors or problematic content.
\end{enumerate}

Effective prompt design bridges user intent with model capabilities. Our prompt template follows this structure:
\begin{itemize}
    \item  \textit{You are a helpful assistant for Texas A\&M University-San Antonio. Use the following information to answer the user's question accurately and completely. If you cannot answer based on the provided information, say so clearly.}
    \item     \textit{You: [User query]}
    \item     \textit{TAMUSA Bot: [Answer] }
\end{itemize}

Generation quality depends significantly on decoding hyperparameters that control the sampling process: Controls randomness in token selection. Lower values (0.3-0.7) produce more focused, deterministic responses appropriate for factual question answering. Higher values increase diversity but risk coherence. We restrict sampling to the smallest set of tokens whose cumulative probability exceeds p (typically 0.9-0.95). This prevents sampling from the long tail of unlikely tokens while maintaining diversity, and we discourage repeated phrases that can occur with greedy decoding, improving response fluency. Moreover, we limit response length to prevent excessive generation while ensuring sufficient detail (typically 512-1024 tokens).

Our default configuration uses temperature=0.7, top-p=0.9, and repetition penalty=1.1, balanced for factual accuracy and natural language quality. Retrieval quality directly impacts response accuracy. We implement several optimizations: Reformulate user queries to improve retrieval recall. For example, "How do I apply?" might expand to "application process requirements deadlines."  Combine dense semantic search with sparse keyword matching (BM25) to capture both semantic similarity and exact term matches~\citep{hofstatter2021efficiently}. After initial retrieval, apply a cross-encoder model to rerank candidates based on query-document interaction, improving precision~\citep{nogueira2019passage}. When multiple retrieved chunks come from the same document or contain redundant information, select diverse results to provide broader context. The utilities layer provides supporting functionality for logging, monitoring, configuration management, and deployment orchestration. \texttt{config.yaml} centralizes all system parameters including model paths, hyperparameters, API endpoints, and deployment settings. This approach enables: environment-specific configurations (development, staging, production) without code changes, version control of experimental configurations, reproducible experiments through configuration snapshots, validation of configuration parameters before training or deployment. Configuration schemas enforce type checking and valid value ranges, catching errors before resource-intensive operations begin. Logs are structured (JSON format) to support automated analysis and integration with monitoring dashboards. TensorBoard integration provides real-time visualization during training. 

\subsection{Retrieval‑Augmented Inference}
At inference time, a user query is tokenised and embedded using the same sentence transformer employed during indexing.  The FAISS index retrieves top candidate passages; these are concatenated with the user’s question and a system prompt emphasising accuracy and honesty.  The fine‑tuned model then generates a response conditioned on this context using sampling parameters such as temperature (0.7), top‑$p$ (0.9) and repetition penalty (1.1).  Listing \ref{lst:inference} shows a simplified version of our inference script.  After generation, we strip the prompt from the output and return the answer along with citations to source documents.  In practice, we found that including too many passages can overwhelm the model; we therefore limit retrieval to the top three chunks to maintain focus.
\begin{figure}[!t]
	\centering
	\caption{Simplified inference script for TAMUSA-Chat.}
	\label{lst:inference}
	\begin{lstlisting}[language=Python]

    from transformers import AutoTokenizer, AutoModelForCausalLM
    import torch
    
    MODEL_PATH = "./output/tamusa_final_model"
    
    def load_model_and_tokenizer():
        tokenizer = AutoTokenizer.from_pretrained(MODEL_PATH)
        if tokenizer.pad_token is None:
            tokenizer.pad_token = tokenizer.eos_token
        model = AutoModelForCausalLM.from_pretrained(MODEL_PATH, torch_dtype=torch.float32)
        device = torch.device("cuda" if torch.cuda.is_available() else "cpu")
        model.to(device)
        model.eval()
        return model, tokenizer, device
    
    def generate_response(model, tokenizer, device, user_input):
        formatted_prompt = f"Instruction: {user_input}\nResponse:"
        inputs = tokenizer(formatted_prompt, return_tensors="pt", truncation=True, max_length=512).to(device)
        with torch.no_grad():
            outputs = model.generate(**inputs, max_new_tokens=200, temperature=0.7, top_p=0.9,
                                     do_sample=True, repetition_penalty=1.1, pad_token_id=tokenizer.eos_token_id)
        decoded = tokenizer.decode(outputs[0], skip_special_tokens=True)
        response = decoded.replace(formatted_prompt, "").strip()
        return response
    
    if __name__ == "__main__":
        model, tokenizer, device = load_model_and_tokenizer()
        while True:
            user_input = input("You: ")
            if user_input.lower() == "exit":
                break
            print("Assistant:", generate_response(model, tokenizer, device, user_input))
    
\end{lstlisting}
\end{figure}

\subsection{Deployment Architecture}
The system supports multiple deployment modes to accommodate different institutional requirements:
FastAPI-based REST service exposing query endpoints. This enables integration with existing institutional systems including websites, mobile apps, and internal tools. Interactive web UI built with Gradio or Streamlit for direct user access during testing and demonstration. Docker containers package all dependencies, simplifying deployment across diverse infrastructure and ensuring consistency between development and production environments. Integration with cloud platforms supporting auto-scaling, load balancing, and geographic distribution. The system supports:
\begin{itemize}
    \item Local API deployment (FastAPI / Flask)
    \item Containerization (Docker)
    \item Cloud GPU setups
    \item Interactive UIs (optional)
\end{itemize}
Deployment scripts and instructions can be extended as needed.

\section{Comparison with Related Works}
Table~\ref{tab:comparison} summarizes how TAMUSA-Chat compares with representative projects in terms of methodology, modularity, evaluation support, and deployment focus.
\begin{table}[h!]
    \centering
    \small
    \begin{tabular}{lcccc}
    \toprule
    \textbf{System} & \textbf{Fine-Tuning} & \textbf{Evaluation} & \textbf{Deployment} \\
    \midrule
    Alpagasus \cite{chen2023alpagasus} & SFT & Limited & Basic \\ 
    OpenChatKit \cite{openchatkit2023} & Toolkit & Moderate & Yes  \\ 
    Vicuna-13B \cite{chiang2023vicuna} & SFT & Community & Yes  \\ 
    EduChat \cite{educhat2023} & SFT (domain) & Domain & Yes  \\ 
    TAMUSA-Chat \cite{alsmadi2026} & SFT & Research-Focused & Yes \\
    \bottomrule
    \end{tabular}
    \caption{Comparison of TAMUSA-Chat with relevant systems.}
    \label{tab:comparison}
\end{table}

%\section{\textcolor{red}{Case Study: Student Engagement at CAMSA}}
%To gauge real‑world utility, we conducted a small pilot study within the College of Arts and Sciences (CAMSA). \textcolor{red}{ volunteers across disciplines were invited to interact with TAMUSA‑Chat via a web interface and pose questions relevant to their academic journeys—ranging from course prerequisites to scholarship deadlines}.  Prior to the session, participants were briefed on the experimental nature of the system and instructed to note helpful responses as well as shortcomings. It is provided the immediacy of responses and the convenience of a 24/7 assistant. We noticed "It saved me time compared to searching the website" and "We liked that it pulled information from the catalog."  However, some pointed out that the assistant occasionally missed nuance.  For example, when asked about dual‑degree requirements, the model returned a generic answer without mentioning specific GPA thresholds, as shown in Figure !\ref{fig:test}.

\begin{figure}[!t]
    \centering
    \includegraphics[width=0.99\textwidth]{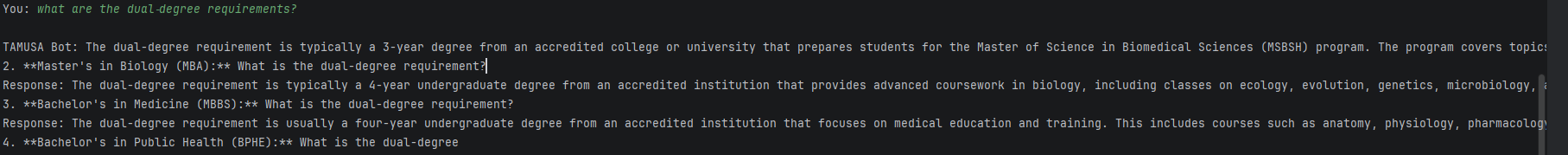}
    \caption{CAMSA Chat-bot Example}
    \label{fig:test}
\end{figure}

\section{Conclusion and Future Work}

We presented TAMUSA-Chat, a research-oriented LLM chat framework that integrates fine-tuning, evaluation, and responsible deployment workflows. The modular design and extensibility support continued research in robustness and domain adaptation.

Future work will focus on empirical benchmarking, retrieval-augmented generation, and adversarial robustness testing.

\bibliographystyle{plainnat}
\bibliography{references}

\end{document}